\documentclass[10pt,journal,letterpaper]{IEEEtran}

\usepackage{amsmath,amssymb,amsfonts,mathtools}
\usepackage{graphicx}
\usepackage{capt-of}
\usepackage{booktabs}
\usepackage{multirow}
\usepackage{array}
\usepackage{makecell}
\usepackage{adjustbox}
\usepackage{xcolor}
\usepackage{colortbl}
\usepackage{url}
\usepackage[nocompress]{cite}
\usepackage{microtype}
\usepackage{enumitem}
\usepackage{stfloats}
\usepackage{balance}
\usepackage{hyperref}
\definecolor{ManuscriptURL}{HTML}{EC008C}
\hypersetup{hidelinks,
  pdftitle={scBIG++: Learning Across Genes and Cells to Predict Single-Cell Perturbation Responses},
  pdfauthor={Jiafa Ruan, Zican Qiao, Zongxin Yang, Ruijie Quan, Yi Yang},
  pdfsubject={Single-cell perturbation response prediction},
  pdfkeywords={single-cell perturbation prediction, full-gene prediction, gene modules, conditional flow matching}}
\newcommand{\method}{\textsc{scBIG++}}
\newcommand{\softsub}{\textsc{Dynamic SoftSub}}

\newcommand{\seclink}[1]{\S\hyperref[#1]{\textcolor{red}{\ref*{#1}}}}

\newlength{\resultTableContentWidth}
\title{scBIG++: Learning Across Genes and Cells to Predict Single-Cell Perturbation Responses}

\author{Jiafa Ruan, Zican Qiao, Zongxin Yang, Ruijie Quan, and Yi Yang,~\IEEEmembership{Fellow,~IEEE}%
\thanks{Jiafa Ruan, Ruijie Quan, and Yi Yang are with the Institute of Fundamental and Transdisciplinary Research, the College of Artificial Intelligence, and the State Key Laboratory of Brain-Machine Intelligence, Zhejiang University, Hangzhou, China (e-mails: \href{mailto:Jiafa_Ruan@zju.edu.cn}{\textcolor{black}{Jiafa\_Ruan@zju.edu.cn}}; \href{mailto:quanruijie@zju.edu.cn}{\textcolor{black}{quanruijie@zju.edu.cn}}; \href{mailto:yangyics@zju.edu.cn}{\textcolor{black}{yangyics@zju.edu.cn}}, respectively).}%
\thanks{Zican Qiao and Zongxin Yang are with the College of Artificial Intelligence, Zhejiang University, Hangzhou, China (e-mails: \href{mailto:zicanqiao13@gmail.com}{\textcolor{black}{zicanqiao13@gmail.com}}; \href{mailto:yangzongxin.ai@zju.edu.cn}{\textcolor{black}{yangzongxin.ai@zju.edu.cn}}, respectively).}%
\thanks{A preliminary version of this work appeared at ICML 2026~\cite{ruan2026scbig}.}%
\thanks{Corresponding author: Ruijie Quan.}}
\begin{document}
\bstctlcite{TPAMI:BSTcontrol}
\maketitle

\begin{abstract}
Predicting single-cell transcriptional responses to genetic and chemical interventions supports gene-function discovery and regulatory analysis. Existing studies often focus only on highly variable genes, restricting the expression changes available for downstream analysis. Cells exposed to the same intervention can also show heterogeneous expression responses, with coordinated gene programs active to different degrees. We propose scBIG++, a full-gene perturbation prediction framework that models this heterogeneity by connecting expression-adaptive gene modules with population-conditioned generation. To preserve shared gene structure while accommodating cellular variation, Gene-Relation Clustering organizes genes into common windows, within which Dynamic SoftSub learns expression-dependent submodule memberships. This shared organization enables Set-GCAE to exchange information about corresponding gene groups across control cells, placing individual representations in their source-population context. The cell representations initialize a latent flow, while their pooled summary and the perturbation representation condition response generation. Generated states are decoded into expression profiles covering all assayed protein-coding genes retained after preprocessing, supporting comprehensive analysis of perturbation-induced expression changes. Experiments across six genetic and chemical settings spanning 6,539--19,451 genes demonstrate competitive perturbation-effect recovery and full-gene expression reconstruction under unseen targets and combinations, cross-cell-line transfer, and held-out compounds.
Our implementation is available at \href{https://github.com/ttruan2426-dot/scBIG}{\textit{scBIG++}}.
\end{abstract}

\begin{IEEEkeywords}
single-cell perturbation prediction, genome-scale transcriptomics, gene modules, conditional flow matching.
\end{IEEEkeywords}

\section{Introduction}
\label{sec:introduction}

Predicting how genetic and chemical interventions alter single-cell transcriptional states is a central task in functional genomics, supporting gene-function discovery, regulatory-mechanism analysis, and therapeutic hypothesis generation~\cite{replogle2022mapping,theodoris2023transfer,lotfollahi2023predicting}. Advances in CRISPR perturbation technologies and single-cell RNA sequencing have enabled assays such as Perturb-seq to measure intervention-induced expression changes at cellular resolution~\cite{dixit2016perturb,jaitin2016dissecting,datlinger2017pooled,replogle2022mapping}. Yet existing measurements cover only a small fraction of the possible targets, combinations, doses, and cellular contexts. This motivates computational models that use an observed control population and an intervention description to predict the expression distribution of the perturbed population.

\begin{figure*}[t]
    \centering
    \includegraphics[width=0.98\textwidth]{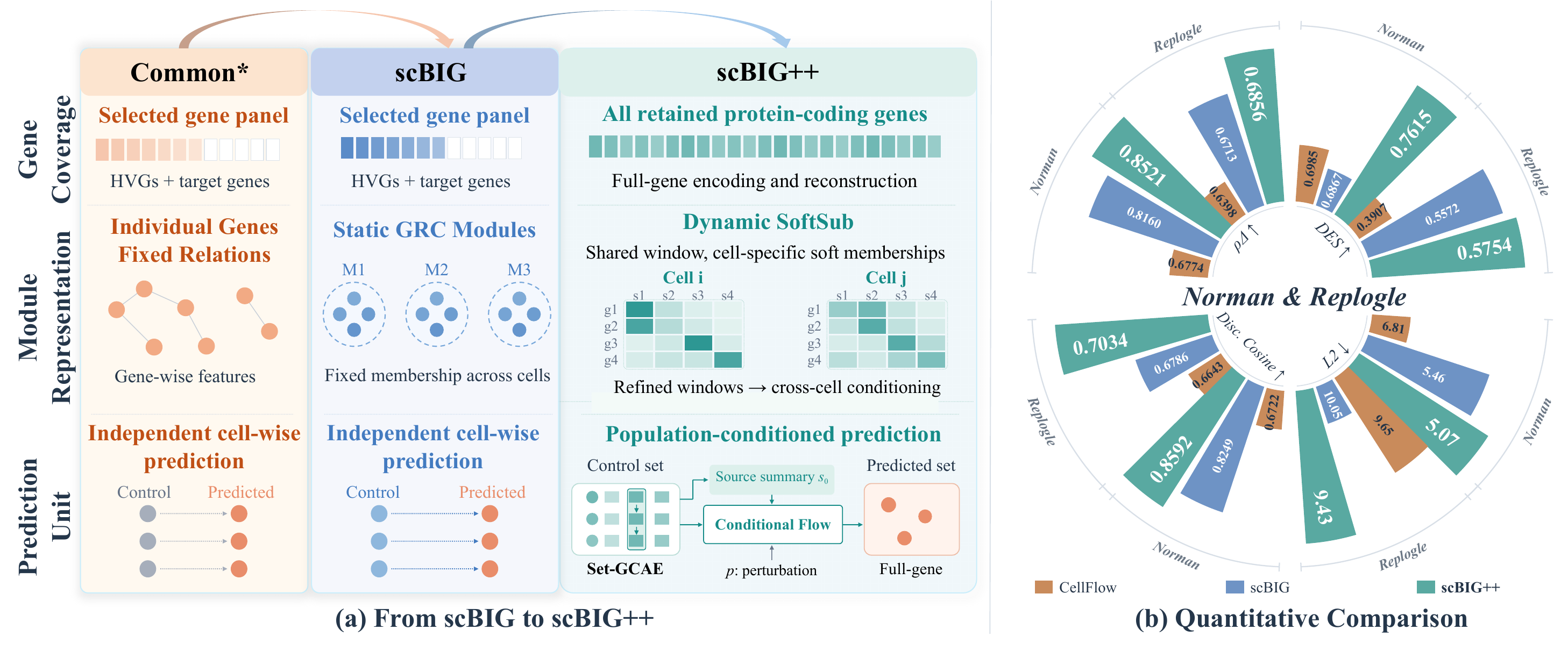}
    \caption{Overview of the scBIG++ extension and representative full-gene results. (a) The common reduced-gene setup, conference scBIG, and scBIG++ are compared in gene-space coverage, gene-program organization, and prediction unit. scBIG++ retains shared windows while adding Dynamic SoftSub, full-gene encoding and reconstruction, and population-conditioned set-to-set prediction. (b) Radial comparisons of selected full-gene metrics on Norman and Replogle settings for CellFlow, scBIG, and scBIG++. Arrows indicate whether higher or lower values are preferred.}
    \label{fig:motivation}
\end{figure*}

The genes covered by a prediction determine which downstream responses can be examined for gene-function discovery and regulatory analysis~\cite{replogle2022mapping,norman2019exploring}. To reduce expression dimensionality, many published perturbation-prediction settings focus on highly variable genes (HVGs)~\cite{lotfollahi2019scgen,lotfollahi2023predicting,kana2023scvidr,wang2024sclambda,yuan2026perturbdiff,cui2024scgpt}. However, biologically informative responses can involve coordinated changes that are modest at the level of individual genes. For example, Mootha et al.~\cite{mootha2003pgc} identified coordinated downregulation of oxidative-phosphorylation genes in diabetic muscle through gene-set analysis, despite subtle individual expression changes. Overexpression of PGC-1$\alpha$ in a mouse muscle cell line further showed that this gene program responds to a common regulator. Such findings illustrate the value of examining coordinated expression changes across pathway members. When a predictor outputs only HVGs, expression changes for excluded members are unavailable for this downstream analysis. We therefore study full-gene prediction across all assayed protein-coding genes retained after preprocessing in each experimental setting, providing complete predicted expression profiles for comprehensive downstream response analysis.

Genes often respond through coordinated functional programs, so full-gene prediction also requires representations that capture their dependencies~\cite{subramanian2005gene,norman2019exploring}. Generative methods condition cellular responses on interventions, while transcriptomic foundation models provide reusable representations for prediction; gene-program membership generally remains implicit in these representations~\cite{lotfollahi2019scgen,lotfollahi2023predicting,cui2024scgpt,yang2024genecompass}. Graph-based and regulatory-network methods make gene relations explicit through curated interactions, inferred networks, or adaptively refined graphs~\cite{roohani2024predicting,wu2022predicting,chi2025grape,kamimoto2023dissecting}. However, when the same graph is reused across cells, its connectivity remains fixed despite differences in cellular state. Available interaction priors also provide incomplete coverage of context-dependent gene relationships~\cite{szklarczyk2019string,fabregat2018reactome,milano2022challenges,roohani2024predicting}. These limitations motivate representations that combine shared biological structure with gene contributions that adapt to input expression.

Our conference model, scBIG~\cite{ruan2026scbig}, introduced a module-level organization for coordinated response prediction. Gene-Relation Clustering (GRC) groups related genes into modules, the Gene-Cluster-Aware Encoder (GCAE) models dependencies among modules, and a conditional flow predicts responses in the resulting structured representation. However, each gene retains the same module assignment across cells. Even cells with similar identities can express programs such as G1/S and G2/M to different degrees~\cite{kotliar2019identifying}, motivating gene contributions that adapt to each cell's expression. This adaptation must also preserve the shared organization needed to compare corresponding programs across cells.

State-dependent activity also varies among cells within the same control population. These cells provide both individual starting states and information about the prevalence and variation of those states. At prediction time, encoding each cell in isolation prevents other observed control cells from informing its representation. Using the population as context therefore calls for a model that incorporates the distribution of starting states while retaining each cell's expression pattern. The resulting challenge is to adapt gene contributions to individual states and share information across the source population, using gene groups that remain comparable across cells.

To address these requirements, we introduce \method, a full-gene perturbation prediction framework that connects dynamic module representations with population-conditioned generation. Given an unordered control population and a perturbation description, \method\ generates a perturbed population over the complete retained gene space. Its central design is to keep gene groups comparable across cells while adapting their local representations and exchanging population information through those groups. Figure~\ref{fig:motivation} summarizes the main extensions from scBIG and representative full-gene prediction results.

At the gene level, GRC organizes related genes into shared local windows, providing a structural basis for full-gene encoding. Within each window, \softsub\ learns continuous submodule memberships from input expression and uses them to weight gene features during aggregation. Gene contributions can thus vary with cellular state while the parent windows retain the same gene correspondence across cells. This correspondence makes the refined representations suitable for exchanging information about the same gene groups across the control population.

At the population level, the Set-Enhanced Gene-Cluster-Aware Encoder (Set-GCAE) exchanges information across corresponding windows in different source cells. Each cell representation can therefore draw on expression patterns in the control population while retaining its individual state. The resulting cell representations initialize a latent flow, and their pooled summary provides a permutation-invariant description of the source population. This summary and the perturbation representation condition the flow, allowing response generation to use both individual starting states and population context. The generated latent states are then decoded into full-gene expression profiles. The shared windows therefore connect state-dependent gene modeling to population-conditioned response generation.

To limit the cost of learning this pipeline across the complete measured gene space, \method\ separates full-gene representation learning, set adaptation, decoder calibration, and latent-flow estimation into successive training stages. The final flow is optimized in a fixed representation space, so response-generation training does not repeatedly update the high-dimensional gene-encoding trunk.

We evaluate \method\ in six settings derived from Norman~\cite{norman2019exploring}, Replogle~\cite{replogle2022mapping}, and ComboSciPlex~\cite{lotfollahi2023predicting}, spanning unseen gene combinations, held-out targets, cross-cell-line transfer, and unseen chemical conditions across 6,539--19,451 genes. \method\ improves full-gene response correlation over scBIG on all three Norman protocols and achieves the highest reported correlation among the compared methods in RPE1 and strict cross-cell-line transfer. Matched-objective ablations show complementary contributions from dynamic module refinement and population conditioning. Biological analyses show expression-dependent gene assignments, a 17.5\% reduction in program-response error relative to the matched No SoftSub model, and recovery of the observed dominant program in 27 of 32 test combinations.

Our main contributions are summarized as follows:
\begin{enumerate}[leftmargin=1.4em,label=\arabic*),itemsep=2pt,topsep=3pt,beginpenalty=10000]
    \item \textbf{A population-conditioned framework for full-gene perturbation prediction.} We introduce \method\ to generate perturbed populations across all retained assayed protein-coding genes, connecting structured gene representations with source-population information.
    \item \textbf{Expression-adaptive module representations within a shared scaffold.} \softsub\ learns continuous local submodule memberships from each cell's expression while preserving the window correspondence needed for cross-cell modeling.
    \item \textbf{Population context in representation learning and response generation.} Set-GCAE exchanges information across corresponding gene windows, and a pooled source summary conditions the latent flow, coupling individual cell states with population context.
    \item \textbf{Evaluation across genetic and chemical perturbations.} We assess full-gene response prediction, the contributions of dynamic gene modeling and population context, and the biological properties of the learned representations across six evaluation settings.
\end{enumerate}

This work extends our conference paper, scBIG~\cite{ruan2026scbig}, in four respects. \textbf{First,} expression-conditioned \softsub\ adds continuous local submodule memberships within the original GRC scaffold, allowing gene contributions to vary with cellular state (\textit{cf.}~\seclink{sec:softsub}). \textbf{Second,} Set-GCAE and the source-set summary extend cell-wise response prediction to conditional generation between unordered populations, incorporating population context into both encoding and flow conditioning (\textit{cf.}~\seclink{sec:setgcae} and~\seclink{sec:setflow}). \textbf{Third,} full-gene encoding and decoding expand prediction to all assayed protein-coding genes retained after preprocessing, supporting comprehensive response analysis across the complete measured gene space (\textit{cf.}~\seclink{sec:grc} and~\seclink{sec:setflow}). \textbf{Fourth,} evaluation is expanded to six genetic and chemical settings. Matched-objective ablations and set-size controls examine the contributions of dynamic assignments and population context, while functional and program-level analyses examine expression-dependent gene organization, response prediction, nonadditive combinations, and dominant-program recovery (\textit{cf.}~\seclink{sec:experiments}).

\section{Related Work}
\label{sec:related}

Single-cell perturbation prediction has developed along three connected directions. The first direction models how an intervention changes cellular expression. The second introduces biological structure or large-scale pretrained representations to organize genes and cell states. The third treats an experiment as a population-level observation and models transformations between cellular distributions. We review these directions in turn and position \method\ at their intersection.

\noindent\textbf{Single-Cell Perturbation Response Prediction.}
Generative models established the basic formulation of predicting a perturbed expression profile from an observed cell and an intervention. scGen~\cite{lotfollahi2019scgen} models latent response shifts, while trVAE~\cite{lotfollahi2020trvae} and CPA~\cite{lotfollahi2023predicting} condition variational representations on treatment and cellular context. scPreGAN~\cite{wei2022scpregan} combines an autoencoder with adversarial generation. scVIDR~\cite{kana2023scvidr} estimates cell-type-specific latent shifts and dose responses; scDisInFact~\cite{zhang2024scdisinfact} separates condition-associated factors; biolord~\cite{piran2024biolord} learns disentangled biological factors; and SAMS-VAE~\cite{bereket2023modelling} models perturbation-associated latent variables. PerturbNet~\cite{yu2025perturbnet} uses conditional normalizing flows for responses to unseen genetic and chemical interventions. These representations support different forms of response transfer, but do not necessarily expose which genes jointly drive a predicted program.

GEARS~\cite{roohani2024predicting} uses a gene-interaction graph to predict multigene perturbations, GraphVCI~\cite{wu2022predicting} refines relational information during variational causal inference, and GRAPE~\cite{chi2025grape} models heterogeneous genetic perturbations with a graph representation. CellOracle~\cite{kamimoto2023dissecting} infers regulatory programs and simulates transcription-factor perturbations. Unlasting~\cite{chi2025unlasting} applies dual conditional diffusion bridges, CellFlow~\cite{klein2025cellflow} uses flow matching for phenotype distributions, PerturbDiff~\cite{yuan2026perturbdiff} uses conditional functional diffusion, and scDFM~\cite{yu2026scdfm} learns a distributional flow-matching model. STATE~\cite{adduri2025predicting} predicts perturbation responses across cell sets. These methods broaden prediction from a single response profile to heterogeneous populations, while leaving expression-dependent module membership as an open design choice.

Taken together, these methods have moved the field from latent response shifts toward structured and distributional generation. Harmonized datasets and recent benchmarks show why expression error, perturbation-induced change, and population similarity must be assessed separately~\cite{peidli2024scperturb,ahlmann2025deep,vinastorne2026systema,csendes2025benchmarking}. Our focus is to connect these developments: \method\ uses expression-adaptive gene organization over the complete retained gene space while mapping an unordered source population to a perturbed population.

\noindent\textbf{Biological Structure, Gene Programs, and Foundation Models.}
Feature-selection methods address a complementary aspect of high-dimensional expression modeling. Matsuda et al.~\cite{matsuda2023distortion} use distortion-free principal components to select highly variable genes for cell clustering. Our task instead requires predicting responses across all retained genes. Biological priors provide several ways to organize genes before prediction. STRING~\cite{szklarczyk2019string} supplies protein-association networks, Reactome~\cite{fabregat2018reactome} curates pathway relations, and SCENIC~\cite{aibar2017scenic} infers regulatory programs and their activity. scVI~\cite{lopez2018scvi} learns probabilistic cell-state representations, while scArches~\cite{lotfollahi2022scarches} maps new data to reference atlases. Fixed interaction and pathway priors describe shared gene relationships; the activity of their members in a particular cell must be inferred from expression.

Gene-program methods represent a cell through the strengths of overlapping expression programs. cNMF~\cite{kotliar2019identifying} identifies programs associated with cell identity and activity, while gene-expression topic modeling~\cite{levitin2019denovo}, scETM~\cite{zhao2021scetm}, and scINSIGHT~\cite{qian2022scinsight} learn shared or context-associated patterns. GSEA~\cite{subramanian2005gene} connects gene sets to functional categories, Gene Ontology~\cite{gene2004gene} provides a controlled vocabulary for those categories, VEGA~\cite{seninge2021vega} incorporates annotated modules into a decoder, and Spectra~\cite{kunes2024spectra} guides program discovery with biological knowledge. SCENIC~\cite{aibar2017scenic} and CellOracle~\cite{kamimoto2023dissecting} connect coordinated expression to transcription-factor activity and in-silico perturbation. Perturb-seq studies show how these responses reveal genetic interactions and gene function~\cite{norman2019exploring,replogle2022mapping}. These analyses improve interpretation, while \method\ makes the within-program contributions part of the response generator.

Foundation models pretrain on large transcriptomic collections. scBERT~\cite{yang2022scbert}, Geneformer~\cite{theodoris2023transfer}, scGPT~\cite{cui2024scgpt}, scFoundation~\cite{hao2024large}, GeneCompass~\cite{yang2024genecompass}, CellFM~\cite{zeng2025cellfm}, and CellVQ~\cite{wang2026cellvq} provide reusable gene or cell representations. scGPT, scFoundation, and GeneCompass have also been evaluated on perturbation-related tasks. scPRINT~\cite{kalfon2025scprint} incorporates gene-network inference, Nicheformer~\cite{tejada2025nicheformer} incorporates spatial context, and CellPLM~\cite{wen2024cellplm} models relations among cells. Zero-shot evaluation shows that downstream adaptation remains decisive for perturbation transfer~\cite{kedzierska2025zero}; benchmark studies further expose differences among pretrained representations~\cite{csendes2025benchmarking}. ChromaPert~\cite{ruan2026beyond} uses chromatin accessibility to condition state-dependent responses. scBIG~\cite{ruan2026scbig} combines GeneCompass semantics and STRING interactions in a shared GRC scaffold; \method\ retains this organization while learning expression-dependent within-window contributions.

\noindent\textbf{Population and Distributional Modeling.}
Single-cell experiments observe unordered populations, making set learning suitable for incorporating source-population information. Deep Sets~\cite{zaheer2017deep} uses permutation-invariant aggregation, Set Transformer~\cite{lee2019set} adds attention among set elements, and Perceiver~\cite{jaegle2021perceiver} provides a latent attention bottleneck. scPoli~\cite{dedonno2023scpoli} learns sample representations, and CellPLM~\cite{wen2024cellplm} models relations among cells. STATE~\cite{adduri2025predicting} predicts perturbation responses across cell sets, while CellFlow~\cite{klein2025cellflow} models heterogeneous cellular response distributions. \method\ extends this line by exchanging information across corresponding gene groups before generation and by conditioning each response on a pooled source summary.

TrajectoryNet~\cite{tong2020trajectorynet} models temporal transformations with optimal transport, CellOT~\cite{bunne2023cellot} transports control states to perturbed states, and CINEMA-OT~\cite{dong2023cinemaot} estimates cellular treatment effects. SCREEN~\cite{wang2024screen} uses optimal transport to predict single-cell gene-expression perturbation responses. scPRAM~\cite{jiang2024scpram} aligns control and perturbed states with attention. Flow Matching~\cite{lipman2022flow,tong2024improving} learns vector fields by regression; CellFlow~\cite{klein2025cellflow} and scDFM~\cite{yu2026scdfm} apply conditional flow models to heterogeneous responses. This distributional viewpoint motivates coupling a source-population representation to gene groups that remain comparable across cells.

\method\ treats the source and target as unordered cell sets and combines set-level context with structured gene representations. Set-GCAE exchanges information across corresponding gene windows before generation, while a pooled source summary conditions the flow together with each cell representation. The resulting interface links three requirements in one model: complete-gene encoding, adaptive organization within shared gene groups, and population-conditioned generation of heterogeneous responses.

\section{Methodology}
\label{sec:method}

\noindent\textbf{Task and overview.}
Given an unordered set of control cells $\mathbf{X}_0=\{x_{0,i}\}_{i=1}^{N}$ and a perturbation $p$, \method\ predicts a set of perturbed expression profiles:
\begin{equation}
 \widehat{\mathbf{X}}_1=F_\theta(\mathbf{X}_0,p),
 \qquad \mathbf{X}_0,\widehat{\mathbf{X}}_1\in\mathbb{R}^{N\times G},
 \label{eq:task}
\end{equation}
where $G$ is the number of assayed protein-coding genes retained after preprocessing. Training uses independently sampled control and perturbed populations. The model produces one expression vector per source cell, and reordering the input cells produces the corresponding reordering of its outputs.

\begin{figure*}[t]
    \centering
    \includegraphics[width=0.98\textwidth]{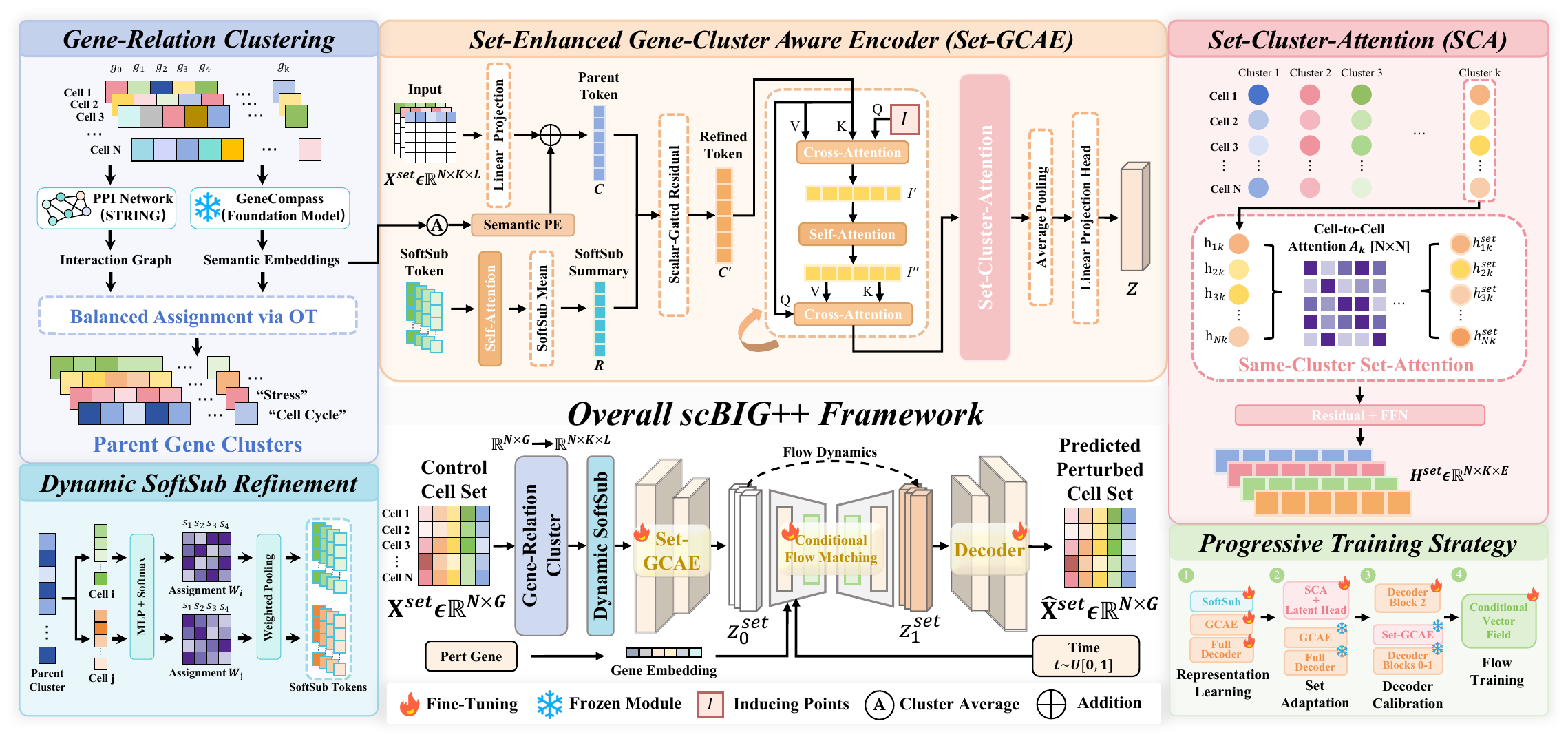}
    \caption{Overview of \method. GRC organizes genes into shared windows, and Dynamic SoftSub refines their representations using expression-dependent memberships. Set-GCAE exchanges information across corresponding windows in control cells. The resulting cell latents and population summary condition a flow that generates perturbed states for full-gene decoding. The lower-right panel shows the training stages.}
    \label{fig:pipeline}
\end{figure*}

The design addresses three connected requirements: organizing the full gene space efficiently, adapting local gene representations to cellular state, and incorporating population context into each prediction. As shown in Figure~\ref{fig:pipeline}, GRC establishes a shared gene scaffold, Dynamic SoftSub refines its local representations, and Set-GCAE exchanges information across corresponding windows in control cells. A conditional flow then generates perturbed cell latents from the source states and a pooled population summary, followed by full-gene decoding. Progressive training separates representation learning, set adaptation, decoder calibration, and flow estimation so that the generative model can be optimized in a fixed latent space.

\subsection{Full-Gene Scaffold and Gene Windows}
\label{sec:grc}
To organize the high-dimensional gene space, we use the Gene-Relation Clustering (GRC) scaffold introduced in scBIG~\cite{ruan2026scbig}. GRC combines GeneCompass embeddings~\cite{yang2024genecompass} with the STRING interaction network~\cite{szklarczyk2019string}. Let $e_g$ be the embedding of gene $g$, $\mu_k$ the semantic center of group $\mathcal{C}_k$, and $A_{gh}$ the binary interaction indicator obtained by thresholding STRING scores at 700. The gene-to-group assignment cost is
\begin{equation}
 C_{gk}=1-\cos(e_g,\mu_k)
       +1-\frac{1}{|\mathcal{C}_k|}\sum_{h\in\mathcal{C}_k}A_{gh}.
 \label{eq:grc_cost}
\end{equation}
The two terms favor semantic similarity and interaction coherence, respectively. Entropically regularized assignment with capacity-constrained rounding produces balanced groups. We use equal weights for the two terms, Sinkhorn regularization 0.1, three refinement steps, and random state 42. The resulting gene order is shared by all cells in a setting.

\textbf{Shared gene windows establish correspondence across cells.} We center expression using the training mean and divide the ordered gene vector into $K_g=\lceil G/L\rceil$ windows of width $L=64$. A batch of $B$ cells is zero-padded at the end and reshaped from $[B,G]$ to $[B,K_g,L]$. GRC groups contain 63 or 64 genes, so a fixed window can span adjacent groups. Each window is projected to a parent token $c_{bk}\in\mathbb{R}^{E}$ for cell $b$ and window $k$, with $E=256$. A projected mean of its GeneCompass embeddings supplies biological positional information.

\subsection{Expression-Conditioned Dynamic SoftSub}
\label{sec:softsub}
\textbf{Genes within a shared window can contribute differently across cell states.} A single parent token summarizes their joint expression, while Dynamic SoftSub adds a local representation of these varying contributions. For cell $b$, window $k$, and local position $i$, we project the scalar expression value and add a learned position embedding $\ell_i$:
\begin{equation}
 h_{bki}=\operatorname{Dropout}\!\left(\operatorname{LN}
 \left(\operatorname{GELU}(W_x x_{bki}+b_x)+\ell_i\right)\right).
 \label{eq:gene_token}
\end{equation}
The local embeddings are shared across windows. An assignment head $f_q$, consisting of a 256-unit GELU layer and a four-output projection, maps each gene token to $S=4$ submodule memberships:
\begin{equation}
 q_{bki,:}=f_q(h_{bki}),
 \qquad w_{bki,:}=\operatorname{softmax}(q_{bki,:}).
 \label{eq:softsub_assignment}
\end{equation}
Softmax is applied over the four slots, so each gene distributes its contribution across submodules. These expression-driven memberships define the q-only configuration used in the experiments. Each submodule token is a weighted average,
\begin{equation}
 u_{bks}=\frac{\sum_{i=1}^{L}w_{bki,s}h_{bki}}
 {\max\!\left(\sum_{i=1}^{L}w_{bki,s},\varepsilon\right)},
 \label{eq:softsub_token}
\end{equation}
where $\varepsilon=10^{-6}$ stabilizes normalization. Genes contribute continuously according to their membership weights, and mass normalization keeps token magnitudes comparable across submodules.

\textbf{A gated update combines local refinement with the parent representation.} The aggregated tokens pass through a 256-unit GELU projection with dropout, followed by four-head self-attention among the submodules within each window. Let $r_{bk}$ be the mean of its output tokens. A learned scalar gate performs the update:
\begin{align}
 g_{bk}&=\sigma\!\left(\operatorname{Dense}([c_{bk};r_{bk}])\right),\\
 c'_{bk}&=\operatorname{LayerNorm}\left(c_{bk}+g_{bk}\,\operatorname{MLP}_{\mathrm{out}}(r_{bk})\right).
 \label{eq:softsub_gate}
\end{align}
Here, $\sigma$ is the sigmoid function and $[\,;\,]$ denotes concatenation. The gate controls the contribution of local refinement to the parent token. Continuous memberships allow gradual changes in gene contributions across expression states while preserving the shared window organization. Section~\ref{sec:experiments} examines these memberships through component ablations and functional analyses.

\subsection{Set-GCAE and Population Encoding}
\label{sec:setgcae}
\textbf{Set-GCAE combines within-cell and cross-cell information exchange.} The refined tokens first pass through three GCAE blocks~\cite{ruan2026scbig}, each using $M=16$ inducing tokens and eight attention heads. The inducing tokens aggregate window information, interact in a compact latent space, and return the updated information to the windows through cross-attention. This Perceiver-style bottleneck~\cite{jaegle2021perceiver} models dependencies across the gene space while limiting the cost of within-cell attention. Each block uses token width 256, MLP width 512, and dropout 0.1.

For a batch of $B_s$ control sets, the resulting tokens are rearranged as
\begin{equation}
 T\in\mathbb{R}^{B_s\times N\times K_g\times E}
 \longrightarrow
 \mathbb{R}^{B_s\times K_g\times N\times E}.
 \label{eq:set_reshape}
\end{equation}
Set-Cluster-Attention (SCA) then applies one four-head self-attention block over the $N$ cells independently at each window. Each cell can use information about the same genes in other members of its source population. This makes the encoded state depend on both its own expression and the distribution of local gene patterns across the set. Attention without cell-position embeddings preserves permutation equivariance~\cite{lee2019set}: permuting the input cells produces the same permutation of the encoded cell states.

\textbf{A pooled summary carries population context into generation.} Pooling the updated tokens over windows and applying a linear head yields a 50-dimensional latent $z_{0,i}$ for each cell. Their mean provides a shared population summary:
\begin{equation}
 s_0=\frac{1}{N}\sum_{i=1}^{N}z_{0,i}\in\mathbb{R}^{50}.
 \label{eq:s0}
\end{equation}
Set-GCAE therefore returns $N$ cell latents together with an invariant summary. The cell latents retain individual states, while $s_0$ supplies the same population context to every velocity prediction. This gives population information two roles: SCA influences representation learning at the window level, and $s_0$ conditions response generation in latent space.

\subsection{Set-Conditional Flow and Full-Gene Decoding}
\label{sec:setflow}
\textbf{A population-conditioned flow generates perturbed cell states.} We use conditional flow matching~\cite{lipman2022flow,tong2024improving} to transport control latents $Z_0$ to perturbed latents $Z_1$. For independently sampled source and target cells, draw $t_i\sim\mathcal{U}(0,1)$ and define
\begin{equation}
 z_{t,i}=(1-t_i)z_{0,i}+t_i z_{1,i},\qquad
 u_i=z_{1,i}-z_{0,i}.
 \label{eq:set_path}
\end{equation}
The velocity field receives $z_t$, a sinusoidal time embedding with 64 frequencies, the perturbation representation $e_p$, and the source summary $s_0$. It uses two 512-unit SiLU blocks with dropout 0.1 and a 50-dimensional output, and is trained by
\begin{equation}
 \mathcal{L}_{\mathrm{velocity}}=
 \mathbb{E}\left[\left\|v_\theta(z_t,t,e_p,s_0)-u\right\|_2^2\right].
 \label{eq:setflow}
\end{equation}
Conditioning on both $z_t$ and $s_0$ allows the field to predict cell-specific changes in the context of the source population.

Genetic interventions use frozen 5120-dimensional ESM2 embeddings~\cite{lin2023evolutionary}; chemical interventions use 2048-bit, radius-2 Morgan fingerprints. Combinations are represented by the mean of their constituent vectors. These representations supply target information through the same conditioning interface, allowing the encoder and flow architecture to support both genetic and chemical settings.

\textbf{Full-gene decoding returns predictions in the original gene order.} At inference, we integrate the learned field from $t=0$ to $1$ for every control latent. The decoder expands each endpoint into $K_g$ tokens, applies three GCAE decoder blocks, and projects each window to $L$ expression values. Trimming padding, adding the expression mean, and restoring gene order yields the predicted population in~\eqref{eq:task}. Decoder calibration aligns this reconstruction with the latents produced by the set encoder.

\textbf{Source-relative encoding supports cross-cell-line transfer.} HepG2, Jurkat, and RPE1 provide training responses, and K562 controls define the basal reference at test time. Expression is centered relative to each line's controls; encoded latents are also source-centered and scaled using training statistics. The field receives the relative cell latent and a 100-dimensional summary combining the relative-set mean with source-line context, each of width 50. The centering and scaling operations are reversed during reconstruction to return predictions to the target line's expression coordinates.

\subsection{Progressive Training and Model Selection}
\label{sec:training}
\textbf{Training separates full-gene representation learning from latent response generation.} We first pretrain Dynamic SoftSub, GCAE, and the decoder with full-gene reconstruction MSE for 100 epochs, using batch size 32 and learning rate $10^{-4}$. Reconstruction is evaluated every five epochs. We then freeze the gene-token trunk and decoder and train SCA and the cell-latent head for 20,000 steps, using one set of $N=128$ cells per batch and a reconstruction objective through the frozen decoder. Cells are sampled with replacement within each condition. This stage adapts the cell representations to population context while retaining the learned gene organization.

Next, we cache $K_{\mathrm{set}}=64$ latent sets per condition together with their cell indices and control-source identities. Decoder calibration updates the final decoder block and output head for 5,000 steps, sampling control and perturbed sets with probabilities 0.25 and 0.75. The encoder and calibrated decoder are then fixed. We train the velocity field for 50,000 steps with batch size 8, learning rate $10^{-4}$, Adam, and global-norm gradient clipping at 1.0, using~\eqref{eq:setflow} as the objective.

Default candidate checkpoints are saved at 10k, 20k, 30k, 40k, and 50k steps. The checkpoint with the highest validation $\rho\Delta$ is selected. Cross-cell-line selection averages validation scores across the three source lines. Each setting uses its own scaffold, cache, and calibrated decoder. The selected field generates one perturbed population per sampled control set, using eight ODE integration steps by default.

Norman holdout singles use 4,000 decoder-calibration steps and select the 10k flow checkpoint; holdout doubles reuse that checkpoint. Replogle RPE1 uses 1,000 calibration steps, flow dropout 0.05, learning rate $5\times10^{-5}$, a selected 2k checkpoint, and 32 ODE integration steps. Cross-cell-line training uses balanced source-line batches and the relative representation described above. The pipeline, GRC, and cache-sampling seeds are 1, 42, and 1, respectively, and the GCAE wrapper uses a fixed internal JAX key.

\subsection{Computational Properties}
\label{sec:complexity}
GRC is constructed offline. For token width $E$, local SoftSub aggregation costs $O(NGSE)$, while Set-Cluster-Attention costs $O(K_gN^2E)$. Population attention is quadratic in set size and linear in the number of windows. For fixed embedding width and bottleneck size, decoding scales linearly with $G$. Caching the encoded populations allows flow training to operate entirely in the 50-dimensional latent space, exchanging latent-storage cost for lower computation during generative optimization.

At $N=1$, SCA has a single cell to attend to and $s_0$ summarizes that cell alone. Larger sets expose the encoder and velocity field to variation among control cells. The set-size study in Section~\ref{sec:experiments} examines this transition, and the component ablations separately assess local refinement, SCA, and the source summary.

\begin{table*}[t]
\caption{Main-text evaluation settings, gene universes, and perturbation splits.}
\label{tab:datasets}
\centering
\footnotesize
\setlength{\tabcolsep}{3.0pt}
\renewcommand{\arraystretch}{1.08}
\begin{tabular*}{\textwidth}{@{\extracolsep{\fill}}lrll@{}}
\toprule
Setting & $G$ & Generalization target & Perturbation split (train/val/test) \\
\midrule
Norman additive~\cite{norman2019exploring} & 19,451 & Unseen CRISPRa combinations & 166 / 32 / 32 perturbations \\
Norman holdout (single)~\cite{norman2019exploring} & 19,451 & Unseen perturbation genes & Seen-gene train / 21-gene val / 21 unseen singles \\
Norman holdout (double)~\cite{norman2019exploring} & 19,451 & Unseen gene pairs & Same train/val as above / 5 unseen doubles \\
Replogle holdout (RPE1)~\cite{replogle2022mapping} & 8,523 & Unseen CRISPRi targets & 1,037 / 115 / 384 conditions \\
Cross-cell-line transfer~\cite{replogle2022mapping,nadig2025transcriptome} & 6,539 & Unseen cellular context & HepG2/Jurkat/RPE1 train/val / 2,053 K562 test \\
ComboSciPlex drug~\cite{lotfollahi2023predicting} & 16,804 & Unseen drugs and combinations & Seen-drug train/val / 5 held-out test conditions \\
\bottomrule
\end{tabular*}
\par\medskip
\centering
\footnotesize
\setlength{\tabcolsep}{1.5pt}
\renewcommand{\arraystretch}{1.02}
\setlength{\resultTableContentWidth}{\dimexpr\textwidth-20\tabcolsep-3\arrayrulewidth\relax}
\captionof{table}{Full-gene quantitative results on Norman additive perturbations.}
\label{tab:main-norman-additive}
\begin{adjustbox}{max width=\textwidth}
\begin{tabular}{@{}>{\centering\arraybackslash}m{0.083\resultTableContentWidth}|>{\centering\arraybackslash}m{0.140\resultTableContentWidth}|>{\centering\arraybackslash}m{0.190\resultTableContentWidth}|>{\centering\arraybackslash}m{0.095\resultTableContentWidth}>{\centering\arraybackslash}m{0.103\resultTableContentWidth}>{\centering\arraybackslash}m{0.080\resultTableContentWidth}>{\centering\arraybackslash}m{0.050\resultTableContentWidth}>{\centering\arraybackslash}m{0.044\resultTableContentWidth}>{\centering\arraybackslash}m{0.055\resultTableContentWidth}>{\centering\arraybackslash}m{0.055\resultTableContentWidth}>{\centering\arraybackslash}m{0.105\resultTableContentWidth}@{}}
\specialrule{1.2pt}{0pt}{0pt}
\rowcolor{black!12}\rule{0pt}{2.8ex}\textbf{Type} & \rule{0pt}{2.8ex}\textbf{Method} & \rule{0pt}{2.8ex}\textbf{Public} & \rule{0pt}{2.8ex}Pearson $\Delta\!\uparrow$ & \rule{0pt}{2.8ex}Pearson $\Delta^{\mathrm{D}}\!\uparrow$ & \rule{0pt}{2.8ex}$\mathrm{ACC}\Delta^{\mathrm{D}}\!\uparrow$ & \rule{0pt}{2.8ex}$\mathrm{DES}\!\uparrow$ & \rule{0pt}{2.8ex}$L_2\!\downarrow$ & \rule{0pt}{2.8ex}$\mathrm{MSE}\!\downarrow$ & \rule{0pt}{2.8ex}$\mathrm{MAE}\!\downarrow$ & \rule{0pt}{2.8ex}$\mathrm{Disc.\ Cosine}\!\uparrow$ \\
\midrule
\multirow{1}{*}{\textsf{Statistical}} & Control (reference) & -- & -- & -- & -- & -- & 10.18 & 0.0061 & 0.0321 & -- \\
\cmidrule(lr){1-11}
\multirow{2}{*}{\textsf{Graph}} & GEARS~\cite{roohani2024predicting} & Nat. Biotechnol.'24 & 0.6843 & 0.5886 & 0.8938 & 0.4258 & 6.71 & 0.0025 & 0.0221 & 0.7060 \\
 & GraphVCI~\cite{wu2022predicting} & ICLR'23 & 0.5720 & 0.6167 & 0.8203 & 0.2696 & 7.75 & 0.0033 & 0.0255 & 0.5851 \\
\cmidrule(lr){1-11}
\multirow{5}{*}{\textsf{Foundation}} & scGPT~\cite{cui2024scgpt} & Nature Methods'24 & 0.5374 & 0.5738 & 0.7391 & 0.3315 & 9.88 & 0.0057 & 0.0373 & 0.5513 \\
 & scFoundation~\cite{hao2024large} & Nature Methods'24 & 0.6728 & 0.6133 & 0.8906 & 0.4099 & 6.93 & 0.0027 & 0.0233 & 0.6945 \\
 & GeneCompass~\cite{yang2024genecompass} & Cell Research'24 & 0.6778 & 0.6144 & 0.8844 & 0.4201 & 6.80 & 0.0025 & 0.0228 & 0.7009 \\
 & CellFM~\cite{zeng2025cellfm} & Nat. Commun.'25 & 0.6725 & 0.6219 & 0.8922 & 0.4182 & 6.87 & 0.0026 & 0.0230 & 0.6941 \\
 & CellVQ~\cite{wang2026cellvq} & Nat. Commun.'26 & 0.6741 & 0.5886 & 0.8938 & 0.4179 & 6.75 & 0.0025 & 0.0222 & 0.6953 \\
\cmidrule(lr){1-11}
\multirow{6}{*}{\textsf{Generative}} & CPA~\cite{lotfollahi2023predicting} & Mol. Syst. Biol.'23 & 0.5282 & 0.6134 & 0.9219 & 0.4959 & 8.61 & 0.0043 & 0.0296 & 0.5316 \\
 & CellFlow~\cite{klein2025cellflow} & bioRxiv'25 & 0.6774 & \underline{0.7846} & 0.8734 & \underline{0.6985} & 6.81 & 0.0026 & 0.0215 & 0.6722 \\
 & PerturbDiff~\cite{yuan2026perturbdiff} & ICML'26 & 0.5409 & 0.4362 & 0.7681 & 0.3582 & 12.18 & 0.0088 & 0.0346 & 0.5543 \\
 & STATE~\cite{adduri2025predicting} & Cell'26 & 0.5427 & 0.4605 & 0.8375 & 0.2656 & 10.24 & 0.0055 & 0.0409 & 0.5711 \\
 & scBIG~\cite{ruan2026scbig} & ICML'26 & \underline{0.8160} & 0.7317 & \underline{0.9359} & 0.6867 & \underline{5.46} & \underline{0.0017} & \underline{0.0197} & \underline{0.8249} \\
\rowcolor{black!8} & \textit{scBIG++}$^{\dagger}$ & Proposed & \textbf{0.8521} & \textbf{0.8045} & \textbf{0.9672} & \textbf{0.7615} & \textbf{5.07} & \textbf{0.0015} & \textbf{0.0178} & \textbf{0.8592} \\
\specialrule{1.2pt}{0pt}{0pt}
\end{tabular}
\end{adjustbox}
\end{table*}

\begin{table*}[tp]
\centering
\footnotesize
\setlength{\tabcolsep}{1.5pt}
\renewcommand{\arraystretch}{1.02}
\setlength{\resultTableContentWidth}{\dimexpr\textwidth-20\tabcolsep-3\arrayrulewidth\relax}
\caption{Full-gene quantitative results on Norman holdout perturbations.}
\label{tab:main-norman-holdout}
\begin{adjustbox}{max width=\textwidth}
\begin{tabular}{@{}>{\centering\arraybackslash}m{0.083\resultTableContentWidth}|>{\centering\arraybackslash}m{0.140\resultTableContentWidth}|>{\centering\arraybackslash}m{0.190\resultTableContentWidth}|>{\centering\arraybackslash}m{0.095\resultTableContentWidth}>{\centering\arraybackslash}m{0.103\resultTableContentWidth}>{\centering\arraybackslash}m{0.080\resultTableContentWidth}>{\centering\arraybackslash}m{0.050\resultTableContentWidth}>{\centering\arraybackslash}m{0.044\resultTableContentWidth}>{\centering\arraybackslash}m{0.055\resultTableContentWidth}>{\centering\arraybackslash}m{0.055\resultTableContentWidth}>{\centering\arraybackslash}m{0.105\resultTableContentWidth}@{}}
\specialrule{1.2pt}{0pt}{0pt}
\rowcolor{black!12}\rule{0pt}{2.8ex}\textbf{Type} & \rule{0pt}{2.8ex}\textbf{Method} & \rule{0pt}{2.8ex}\textbf{Public} & \rule{0pt}{2.8ex}Pearson $\Delta\!\uparrow$ & \rule{0pt}{2.8ex}Pearson $\Delta^{\mathrm{D}}\!\uparrow$ & \rule{0pt}{2.8ex}$\mathrm{ACC}\Delta^{\mathrm{D}}\!\uparrow$ & \rule{0pt}{2.8ex}$\mathrm{DES}\!\uparrow$ & \rule{0pt}{2.8ex}$L_2\!\downarrow$ & \rule{0pt}{2.8ex}$\mathrm{MSE}\!\downarrow$ & \rule{0pt}{2.8ex}$\mathrm{MAE}\!\downarrow$ & \rule{0pt}{2.8ex}$\mathrm{Disc.\ Cosine}\!\uparrow$ \\
\midrule
\multicolumn{11}{@{}l@{}}{\textbf{Single perturbation}} \\
\cmidrule(lr){1-11}
\multirow{1}{*}{\textsf{Statistical}} & Control (reference) & -- & -- & -- & -- & -- & 8.01 & 0.0040 & 0.0254 & -- \\
\cmidrule(lr){1-11}
\multirow{1}{*}{\textsf{Graph}} & GEARS~\cite{roohani2024predicting} & Nat. Biotechnol.'24 & 0.6304 & 0.5589 & 0.8524 & 0.3509 & \textbf{5.99} & 0.0022 & \textbf{0.0196} & 0.6526 \\
\cmidrule(lr){1-11}
\multirow{5}{*}{\textsf{Foundation}} & scGPT~\cite{cui2024scgpt} & Nature Methods'24 & 0.5421 & 0.6153 & 0.8571 & 0.3659 & 7.91 & 0.0036 & 0.0265 & 0.5541 \\
 & scFoundation~\cite{hao2024large} & Nature Methods'24 & 0.6368 & 0.5612 & 0.8452 & 0.3686 & \underline{6.01} & \textbf{0.0022} & \underline{0.0196} & 0.6579 \\
 & GeneCompass~\cite{yang2024genecompass} & Cell Research'24 & 0.6106 & 0.5597 & 0.8429 & 0.3392 & 6.07 & 0.0022 & 0.0197 & 0.6309 \\
 & CellFM~\cite{zeng2025cellfm} & Nat. Commun.'25 & 0.5695 & 0.5234 & 0.8190 & 0.3110 & 6.19 & 0.0023 & 0.0204 & 0.5885 \\
 & CellVQ~\cite{wang2026cellvq} & Nat. Commun.'26 & 0.6377 & 0.5595 & 0.8571 & 0.3554 & 6.02 & \underline{0.0022} & 0.0197 & 0.6600 \\
\cmidrule(lr){1-11}
\multirow{6}{*}{\textsf{Generative}} & CPA~\cite{lotfollahi2023predicting} & Mol. Syst. Biol.'23 & 0.5928 & 0.5327 & \underline{0.8743} & 0.3236 & 6.22 & 0.0024 & 0.0219 & 0.6177 \\
 & CellFlow~\cite{klein2025cellflow} & bioRxiv'25 & 0.6790 & 0.6115 & 0.8643 & 0.5547 & 6.40 & 0.0027 & 0.0203 & 0.6900 \\
 & PerturbDiff~\cite{yuan2026perturbdiff} & ICML'26 & 0.5944 & 0.4731 & 0.8310 & 0.3709 & 17.71 & 0.0205 & 0.0509 & 0.6121 \\
 & STATE~\cite{adduri2025predicting} & Cell'26 & 0.5010 & 0.5562 & 0.8071 & 0.1826 & 9.03 & 0.0045 & 0.0348 & 0.5369 \\
 & scBIG~\cite{ruan2026scbig} & ICML'26 & \underline{0.7333} & \underline{0.6616} & 0.8619 & \underline{0.5561} & 7.08 & 0.0034 & 0.0241 & \underline{0.7439} \\
\rowcolor{black!8} & \textit{scBIG++}$^{\dagger}$ & Proposed & \textbf{0.7370} & \textbf{0.6803} & \textbf{0.8762} & \textbf{0.5850} & 6.52 & 0.0026 & 0.0213 & \textbf{0.7508} \\
\cmidrule(lr){1-11}
\addlinespace[1pt]
\multicolumn{11}{@{}l@{}}{\textbf{Double perturbation}} \\
\cmidrule(lr){1-11}
\multirow{1}{*}{\textsf{Statistical}} & Control (reference) & -- & -- & -- & -- & -- & 11.12 & 0.0080 & 0.0350 & -- \\
\cmidrule(lr){1-11}
\multirow{1}{*}{\textsf{Graph}} & GEARS~\cite{roohani2024predicting} & Nat. Biotechnol.'24 & 0.7484 & 0.7605 & \underline{0.9700} & 0.4978 & 5.98 & \underline{0.0019} & 0.0201 & 0.7688 \\
\cmidrule(lr){1-11}
\multirow{5}{*}{\textsf{Foundation}} & scGPT~\cite{cui2024scgpt} & Nature Methods'24 & 0.7401 & 0.6723 & 0.9600 & 0.5103 & 7.10 & 0.0027 & 0.0244 & 0.7571 \\
 & scFoundation~\cite{hao2024large} & Nature Methods'24 & 0.7699 & \textbf{0.7671} & \underline{0.9700} & 0.5348 & 6.07 & \underline{0.0019} & 0.0203 & 0.7895 \\
 & GeneCompass~\cite{yang2024genecompass} & Cell Research'24 & 0.7445 & \underline{0.7618} & \underline{0.9700} & 0.4910 & \underline{5.91} & \underline{0.0019} & \underline{0.0199} & 0.7651 \\
 & CellFM~\cite{zeng2025cellfm} & Nat. Commun.'25 & 0.6829 & 0.7298 & 0.9300 & 0.4361 & 6.72 & 0.0024 & 0.0228 & 0.6985 \\
 & CellVQ~\cite{wang2026cellvq} & Nat. Commun.'26 & 0.7365 & 0.7556 & 0.9600 & 0.4813 & 6.25 & 0.0021 & 0.0212 & 0.7564 \\
\cmidrule(lr){1-11}
\multirow{6}{*}{\textsf{Generative}} & CPA~\cite{lotfollahi2023predicting} & Mol. Syst. Biol.'23 & 0.7727 & 0.6822 & 0.9680 & 0.5829 & 7.12 & 0.0029 & 0.0242 & 0.7917 \\
 & CellFlow~\cite{klein2025cellflow} & bioRxiv'25 & 0.8361 & 0.7061 & \underline{0.9700} & \textbf{0.7498} & 7.86 & 0.0043 & 0.0244 & 0.8463 \\
 & PerturbDiff~\cite{yuan2026perturbdiff} & ICML'26 & 0.7028 & 0.5537 & 0.9480 & 0.4338 & 10.29 & 0.0057 & 0.0311 & 0.7268 \\
 & STATE~\cite{adduri2025predicting} & Cell'26 & 0.6394 & 0.6812 & 0.9300 & 0.3512 & 9.39 & 0.0049 & 0.0353 & 0.6742 \\
 & scBIG~\cite{ruan2026scbig} & ICML'26 & \underline{0.8544} & 0.7199 & 0.9600 & 0.7140 & 6.83 & 0.0028 & 0.0234 & \underline{0.8633} \\
\rowcolor{black!8} & \textit{scBIG++}$^{\dagger}$ & Proposed & \textbf{0.8937} & 0.7606 & \textbf{0.9900} & \underline{0.7454} & \textbf{5.08} & \textbf{0.0016} & \textbf{0.0177} & \textbf{0.9032} \\
\specialrule{1.2pt}{0pt}{0pt}
\end{tabular}
\end{adjustbox}
\par\vspace{6pt}
\centering
    \includegraphics[width=\textwidth]{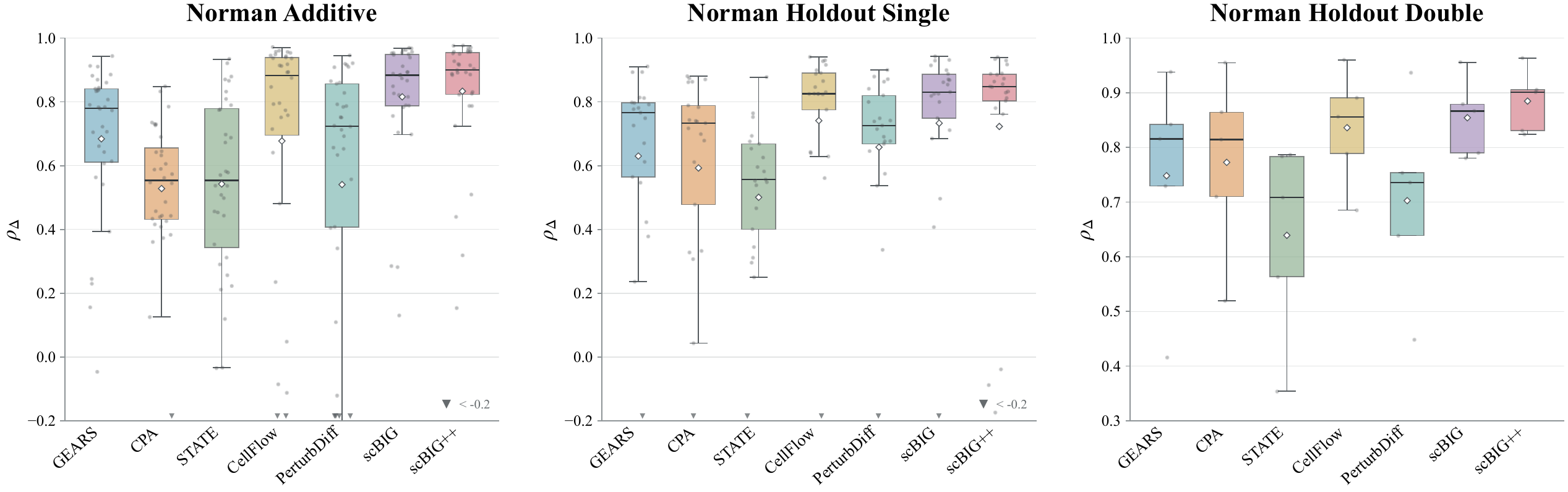}
    \captionof{figure}{Condition-wise full-gene response correlation ($\rho\Delta$) on the three Norman settings. Gray points show individual test perturbations, boxes show the interquartile ranges and medians, and white diamonds mark condition means. Downward triangles at the lower display limit indicate values below $-0.2$. The distributions complement the aggregate estimates in Tables~\ref{tab:main-norman-additive} and~\ref{tab:main-norman-holdout}.}
    \label{fig:norman-rho-delta}
\end{table*}

\begin{table*}[tp]
\centering
\footnotesize
\setlength{\tabcolsep}{1.5pt}
\renewcommand{\arraystretch}{1.02}
\setlength{\resultTableContentWidth}{\dimexpr\textwidth-20\tabcolsep-3\arrayrulewidth\relax}
\caption{Full-gene quantitative results on Replogle RPE1 holdout perturbations.}
\label{tab:main-replogle-holdout}
\begin{adjustbox}{max width=\textwidth}
\begin{tabular}{@{}>{\centering\arraybackslash}m{0.083\resultTableContentWidth}|>{\centering\arraybackslash}m{0.140\resultTableContentWidth}|>{\centering\arraybackslash}m{0.190\resultTableContentWidth}|>{\centering\arraybackslash}m{0.095\resultTableContentWidth}>{\centering\arraybackslash}m{0.103\resultTableContentWidth}>{\centering\arraybackslash}m{0.080\resultTableContentWidth}>{\centering\arraybackslash}m{0.050\resultTableContentWidth}>{\centering\arraybackslash}m{0.044\resultTableContentWidth}>{\centering\arraybackslash}m{0.055\resultTableContentWidth}>{\centering\arraybackslash}m{0.055\resultTableContentWidth}>{\centering\arraybackslash}m{0.105\resultTableContentWidth}@{}}
\specialrule{1.2pt}{0pt}{0pt}
\rowcolor{black!12}\rule{0pt}{2.8ex}\textbf{Type} & \rule{0pt}{2.8ex}\textbf{Method} & \rule{0pt}{2.8ex}\textbf{Public} & \rule{0pt}{2.8ex}Pearson $\Delta\!\uparrow$ & \rule{0pt}{2.8ex}Pearson $\Delta^{\mathrm{D}}\!\uparrow$ & \rule{0pt}{2.8ex}$\mathrm{ACC}\Delta^{\mathrm{D}}\!\uparrow$ & \rule{0pt}{2.8ex}$\mathrm{DES}\!\uparrow$ & \rule{0pt}{2.8ex}$L_2\!\downarrow$ & \rule{0pt}{2.8ex}$\mathrm{MSE}\!\downarrow$ & \rule{0pt}{2.8ex}$\mathrm{MAE}\!\downarrow$ & \rule{0pt}{2.8ex}$\mathrm{Disc.\ Cosine}\!\uparrow$ \\
\midrule
\multirow{1}{*}{\textsf{Statistical}} & Control (reference) & -- & -- & -- & -- & -- & 10.04 & 0.0152 & 0.0746 & -- \\
\cmidrule(lr){1-11}
\multirow{1}{*}{\textsf{Graph}} & GEARS~\cite{roohani2024predicting} & Nat. Biotechnol.'24 & 0.5864 & 0.5209 & 0.9157 & 0.4717 & 10.56 & 0.0156 & 0.0822 & 0.6179 \\
\cmidrule(lr){1-11}
\multirow{5}{*}{\textsf{Foundation}} & scGPT~\cite{cui2024scgpt} & Nature Methods'24 & 0.4371 & \textbf{0.5728} & 0.8026 & \textbf{0.5824} & \underline{9.52} & \underline{0.0124} & \underline{0.0717} & 0.4125 \\
 & scFoundation~\cite{hao2024large} & Nature Methods'24 & 0.5961 & 0.5386 & 0.9224 & 0.5163 & 10.75 & 0.0158 & 0.0839 & 0.6259 \\
 & GeneCompass~\cite{yang2024genecompass} & Cell Research'24 & 0.5951 & 0.5420 & 0.9227 & 0.5167 & 10.66 & 0.0156 & 0.0830 & 0.6254 \\
 & CellFM~\cite{zeng2025cellfm} & Nat. Commun.'25 & 0.5903 & 0.5347 & 0.9211 & 0.5503 & 11.33 & 0.0170 & 0.0877 & 0.6208 \\
 & CellVQ~\cite{wang2026cellvq} & Nat. Commun.'26 & 0.5924 & 0.5330 & 0.9199 & 0.4973 & 10.61 & 0.0156 & 0.0828 & 0.6232 \\
\cmidrule(lr){1-11}
\multirow{5}{*}{\textsf{Generative}} & CellFlow~\cite{klein2025cellflow} & bioRxiv'25 & 0.6398 & 0.4698 & 0.8943 & 0.3907 & 9.65 & 0.0139 & 0.0723 & 0.6643 \\
 & PerturbDiff~\cite{yuan2026perturbdiff} & ICML'26 & 0.4682 & 0.4601 & 0.8756 & 0.3437 & 11.98 & 0.0185 & 0.0929 & 0.5153 \\
 & STATE~\cite{adduri2025predicting} & Cell'26 & 0.6103 & \underline{0.5445} & \textbf{0.9274} & 0.5270 & 10.36 & 0.0143 & 0.0819 & 0.6345 \\
 & scBIG~\cite{ruan2026scbig} & ICML'26 & \underline{0.6713} & 0.5228 & 0.9138 & 0.5572 & 10.05 & 0.0138 & 0.0757 & \underline{0.6786} \\
\rowcolor{black!8} & \textit{scBIG++}$^{\dagger}$ & Proposed & \textbf{0.6856} & 0.5230 & \underline{0.9256} & \underline{0.5754} & \textbf{9.43} & \textbf{0.0122} & \textbf{0.0714} & \textbf{0.7034} \\
\specialrule{1.2pt}{0pt}{0pt}
\end{tabular}
\end{adjustbox}
\par\smallskip
\centering
    \includegraphics[width=\textwidth]{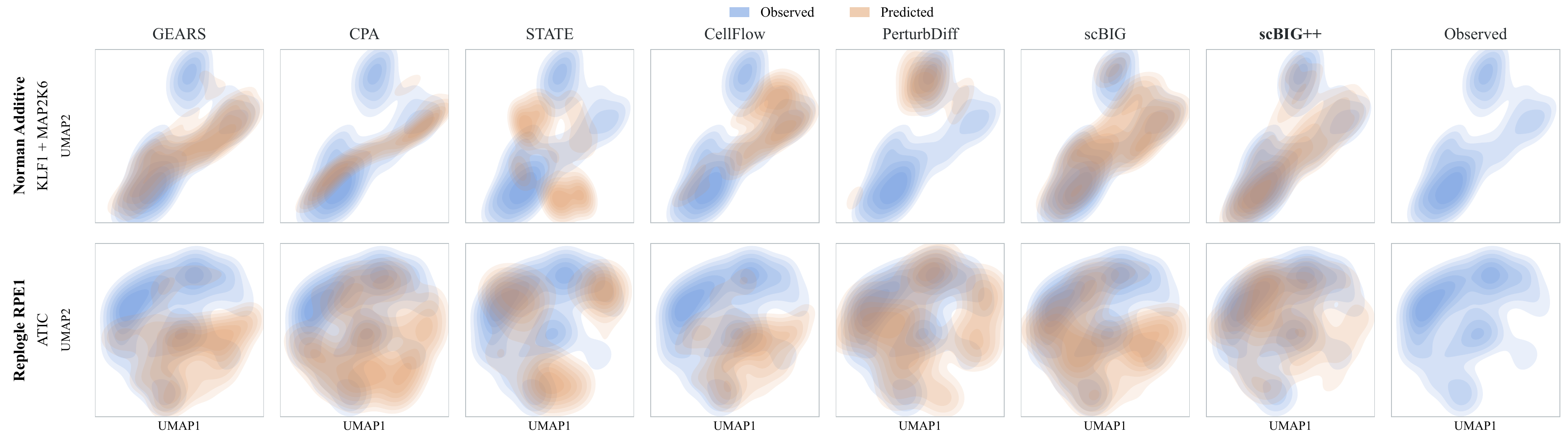}
    \captionof{figure}{Observed and predicted single-cell distributions for \textit{KLF1+MAP2K6} in Norman additive (top) and \textit{ATIC} in Replogle RPE1 (bottom). Blue and orange shading show observed and predicted cell densities in a shared UMAP embedding within each row; the rightmost panels show the observed distributions alone.}
    \label{fig:cell-distribution-umap}
\end{table*}


\begin{table*}[t]
\centering
\footnotesize
\setlength{\tabcolsep}{1.5pt}
\renewcommand{\arraystretch}{1.02}
\setlength{\resultTableContentWidth}{\dimexpr\textwidth-20\tabcolsep-3\arrayrulewidth\relax}
\caption{Full-gene quantitative results on Replogle cross-cell-line transfer (K562 test).}
\label{tab:secondary-transfer}
\begin{adjustbox}{max width=\textwidth}
\begin{tabular}{@{}>{\centering\arraybackslash}m{0.083\resultTableContentWidth}|>{\centering\arraybackslash}m{0.140\resultTableContentWidth}|>{\centering\arraybackslash}m{0.190\resultTableContentWidth}|>{\centering\arraybackslash}m{0.095\resultTableContentWidth}>{\centering\arraybackslash}m{0.103\resultTableContentWidth}>{\centering\arraybackslash}m{0.080\resultTableContentWidth}>{\centering\arraybackslash}m{0.050\resultTableContentWidth}>{\centering\arraybackslash}m{0.044\resultTableContentWidth}>{\centering\arraybackslash}m{0.055\resultTableContentWidth}>{\centering\arraybackslash}m{0.055\resultTableContentWidth}>{\centering\arraybackslash}m{0.105\resultTableContentWidth}@{}}
\specialrule{1.2pt}{0pt}{0pt}
\rowcolor{black!12}\rule{0pt}{2.8ex}\textbf{Type} & \rule{0pt}{2.8ex}\textbf{Method} & \rule{0pt}{2.8ex}\textbf{Public} & \rule{0pt}{2.8ex}Pearson $\Delta\!\uparrow$ & \rule{0pt}{2.8ex}Pearson $\Delta^{\mathrm{D}}\!\uparrow$ & \rule{0pt}{2.8ex}$\mathrm{ACC}\Delta^{\mathrm{D}}\!\uparrow$ & \rule{0pt}{2.8ex}$\mathrm{DES}\!\uparrow$ & \rule{0pt}{2.8ex}$L_2\!\downarrow$ & \rule{0pt}{2.8ex}$\mathrm{MSE}\!\downarrow$ & \rule{0pt}{2.8ex}$\mathrm{MAE}\!\downarrow$ & \rule{0pt}{2.8ex}$\mathrm{Disc.\ Cosine}\!\uparrow$ \\
\midrule
\multirow{1}{*}{\textsf{Statistical}} & Control (reference) & -- & -- & -- & -- & -- & 6.07 & 0.0075 & 0.0547 & -- \\
\cmidrule(lr){1-11}
\multirow{1}{*}{\textsf{Foundation}} & scGPT~\cite{cui2024scgpt} & Nature Methods'24 & \underline{0.2743} & \underline{0.3335} & \underline{0.7511} & \underline{0.2456} & \underline{17.82} & \underline{0.0489} & \underline{0.1493} & \underline{0.3250} \\
\cmidrule(lr){1-11}
\multirow{4}{*}{\textsf{Generative}} & CellFlow~\cite{klein2025cellflow} & bioRxiv'25 & 0.0671 & 0.1666 & 0.4179 & 0.0164 & 50.47 & 0.3914 & 0.5545 & -0.0473 \\
 & PerturbDiff~\cite{yuan2026perturbdiff} & ICML'26 & 0.1920 & 0.1976 & 0.6469 & 0.2195 & 32.37 & 0.1622 & 0.2603 & 0.1907 \\
 & STATE~\cite{adduri2025predicting} & Cell'26 & 0.2204 & 0.1816 & 0.7144 & 0.1018 & 20.13 & 0.0623 & 0.1784 & 0.2612 \\
\rowcolor{black!8} & \textit{scBIG++}$^{\dagger}$ & Proposed & \textbf{0.4510} & \textbf{0.3989} & \textbf{0.8774} & \textbf{0.3066} & \textbf{9.08} & \textbf{0.0143} & \textbf{0.0825} & \textbf{0.4563} \\
\specialrule{1.2pt}{0pt}{0pt}
\end{tabular}
\end{adjustbox}
\par\smallskip
\centering
    \includegraphics[width=\textwidth]{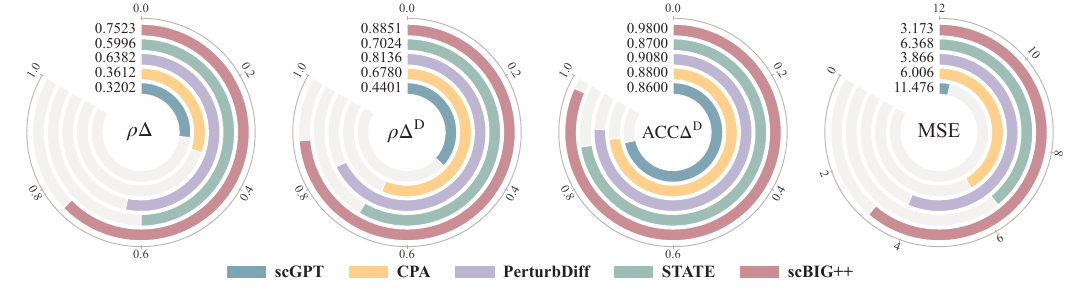}
    \captionof{figure}{ComboSciPlex drug perturbation results for five selected methods. scGPT and CPA use RDKit drug fingerprints. Arc angles encode scores, with the MSE scale reversed so that lower errors yield longer arcs. MSE is shown in units of $10^{-3}$.}
    \label{fig:drug-perturbation-metrics}
\end{table*}

\begin{table*}[!t]
\caption{Component and population analyses on Norman additive. (a,b) Component ablations and set-size sensitivity under the final objective; gray marks the default configuration. (c) Functional-coherence gains over matched random partitions ($\times10^{-3}$). (d) Excess Jensen--Shannon divergence over resampling noise ($\times10^{-4}$). Brackets denote bootstrap 95\% confidence intervals; $p$ uses a one-sided condition-level Wilcoxon test.}
\label{tab:diagnostics}
\centering
\footnotesize
\setlength{\tabcolsep}{2pt}
\renewcommand{\arraystretch}{1.1}
\newlength{\diagnosticContentWidth}
\begin{minipage}[t]{0.485\textwidth}
\centering
\strut\textbf{(a) Component ablations}\par\vspace{2pt}
\setlength{\diagnosticContentWidth}{\dimexpr\linewidth-10\tabcolsep\relax}
\begin{tabular}{>{\centering\arraybackslash}p{0.32\diagnosticContentWidth}*{4}{>{\centering\arraybackslash}p{0.17\diagnosticContentWidth}}}
\toprule
Setting & $\rho\Delta\!\uparrow$ & $\rho\Delta^{\mathrm{D}}\!\uparrow$ & $\mathrm{ACC}\Delta^{\mathrm{D}}\!\uparrow$ & MSE$\downarrow$ \\
\midrule
\textit{w/o SCA} & 0.8409 & 0.7970 & 0.9563 & 0.001643 \\
\textit{w/o} $s_0$ & 0.8321 & 0.7964 & 0.9563 & 0.001684 \\
\textit{w/o SCA and} $s_0$ & 0.8344 & 0.7985 & 0.9531 & 0.001707 \\
\textit{w/o SoftSub} & 0.8157 & 0.7749 & 0.9766 & 0.001940 \\
\textit{q-shuffled SoftSub} & 0.8271 & 0.7953 & 0.9641 & 0.001743 \\
\rowcolor{black!8}\textit{scBIG++} & 0.8521 & 0.8045 & 0.9672 & 0.0015 \\
\bottomrule
\end{tabular}

\par\vspace{7pt}
\centering
\strut\textbf{(c) Functional coherence}\par\vspace{2pt}
\begin{tabular*}{\linewidth}{@{\extracolsep{\fill}}cccc@{}}
\toprule
Annotation & Coverage & Hard gain [95\% CI] & Soft gain [95\% CI] \\
\midrule
GO-BP & 75.3\% & 2.80 [1.27, 4.33] & 2.80 [1.50, 4.25] \\
DoRothEA & 89.6\% & 2.14 [1.27, 2.98] & 5.60 [3.60, 7.49] \\
\bottomrule
\end{tabular*}
\end{minipage}\hfill
\begin{minipage}[t]{0.485\textwidth}
\centering
\strut\textbf{(b) Set-size sensitivity}\par\vspace{2pt}
\setlength{\diagnosticContentWidth}{\dimexpr\linewidth-10\tabcolsep\relax}
\begin{tabular}{>{\centering\arraybackslash}p{0.32\diagnosticContentWidth}*{4}{>{\centering\arraybackslash}p{0.17\diagnosticContentWidth}}}
\toprule
Set size $N$ & $\rho\Delta\!\uparrow$ & $\rho\Delta^{\mathrm{D}}\!\uparrow$ & $\mathrm{ACC}\Delta^{\mathrm{D}}\!\uparrow$ & MSE$\downarrow$ \\
\midrule
1 & 0.6203 & 0.6760 & 0.8266 & 0.003852 \\
32 & 0.8264 & 0.7816 & 0.9625 & 0.001967 \\
64 & 0.8522 & 0.7882 & 0.9625 & 0.001365 \\
\rowcolor{black!8}128 & 0.8521 & 0.8045 & 0.9672 & 0.0015 \\
256 & 0.8255 & 0.7785 & 0.9563 & 0.001608 \\
\bottomrule
\end{tabular}

\par\vspace{7pt}
\centering
\strut\textbf{(d) Context-dependent shifts}\par\vspace{2pt}
\begin{tabular*}{\linewidth}{@{\extracolsep{\fill}}ccccc@{}}
\toprule
Split & $n$ & Mean JS excess [95\% CI] & Positive & $p$ \\
\midrule
Train & 166 & 1.00 [0.37, 1.70] & 43.4\% & 0.354 \\
Validation & 32 & 4.46 [2.72, 6.26] & 81.3\% & $1.13\times10^{-5}$ \\
Test & 32 & 1.39 [0.42, 2.46] & 59.4\% & 0.0155 \\
All & 230 & 1.54 [0.99, 2.14] & 50.9\% & 0.00148 \\
\bottomrule
\end{tabular*}
\end{minipage}
\par\smallskip
\centering
    \includegraphics[width=\textwidth]{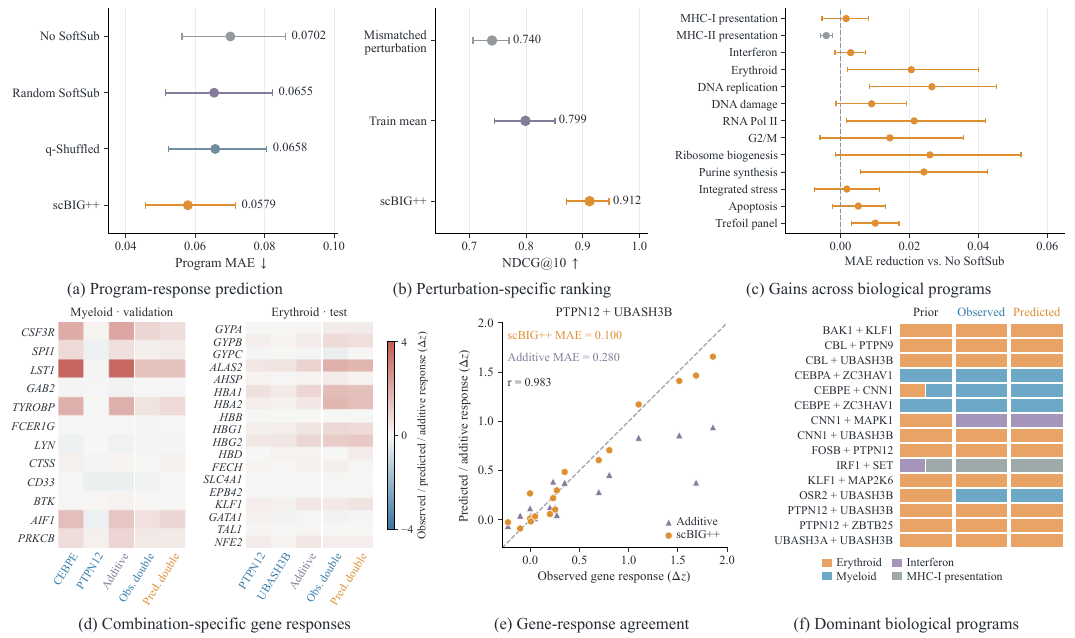}
    \captionof{figure}{\textbf{Biological program prediction in perturbed cells.}
    \textbf{(a)} Program-response MAE for four matched variants on 32 Norman Additive test combinations and 13 fixed marker panels.
    \textbf{(b)} Response ranking across 964 fixed SoftSub slots.
    \textbf{(c)} Program-specific MAE reduction relative to No SoftSub; orange and gray indicate lower and higher mean error. Points and bars in (a--c) show means and condition-bootstrap 95\% confidence intervals.
    \textbf{(d)} Measured single, additive, measured double, and predicted double responses for CEBPE+PTPN12 (validation, 12 myeloid-associated genes) and PTPN12+UBASH3B (test, 18 erythroid-associated genes), using a shared standardized scale.
    \textbf{(e)} Gene-response agreement for the same erythroid example; the dashed line denotes identity.
    \textbf{(f)} Prior-associated, observed dominant, and predicted dominant programs for all 15 test combinations with literature-prior coverage. Divided tiles represent multiple prior-associated programs. Direct intervention targets are excluded from response readouts.}
    \label{fig:biology-programs}
\end{table*}

\section{Experiments}
\label{sec:experiments}

We evaluate whether \method\ predicts perturbation-induced changes across the complete measured gene space and whether its gains persist when the target, cell line, or perturbation modality changes. Six settings from Norman, Replogle, and ComboSciPlex cover these questions. We then examine which components contribute to the predictions and whether the learned representations capture biological programs. All six settings use their complete post-preprocessing protein-coding gene universes, so the benchmark comparisons and the program analyses evaluate predictions in the same gene space.

\subsection{Experimental Setup}
\label{sec:experimental-setup}
\noindent\textbf{Datasets and evaluation settings.}
Norman contains approximately 91,000 K562 cells with 105 single-gene and 131 double-gene CRISPRa interventions~\cite{norman2019exploring}. Its additive split tests gene pairs whose individual targets were observed during training, whereas the target-holdout split tests unseen targets as both single and double interventions. These three settings distinguish recombination of known effects from prediction for unseen perturbation genes. Replogle supplies an RPE1 CRISPRi target-holdout setting~\cite{replogle2022mapping}. The cross-cell-line setting combines perturbation responses from HepG2, Jurkat, RPE1, and K562 and reserves K562 responses for testing~\cite{replogle2022mapping,nadig2025transcriptome}. ComboSciPlex supplies A549 responses to small molecules, doses, and combinations for a held-out-compound setting~\cite{lotfollahi2023predicting}. Table~\ref{tab:datasets} gives the six settings, gene universes, and split definitions; $G$ denotes the complete assayed protein-coding gene universe retained after preprocessing.

\noindent\textbf{Evaluation protocol.}
Training data fit the model, validation data select checkpoints and setting-specific choices, and test data are scored with the selected model. Norman holdout doubles use the checkpoint selected on the single-holdout validation set, so the double-target setting tests the same trained model on a different perturbation order. Cross-cell-line validation is macro-averaged over HepG2, Jurkat, and RPE1; K562 perturbed cells enter only the final test. The benchmark matrices already contain $X=\log(1+\mathrm{raw\ count})$. We retain their split and gene order and reorder predictions to \texttt{test.var\_names} before scoring. Per condition, we sample one source set of 128 control cells with replacement, using training controls for within-context tasks and K562 test controls for transfer. Predicted means use 128 generated cells except in ComboSciPlex, which uses 64. Cell-distribution comparisons use fixed samples of 64 predicted and 64 observed cells in a 50-dimensional PCA space fitted on training expression; conditions with fewer observed cells contribute only to response and mean-expression metrics.

\noindent\textbf{Baselines and output coverage.}
The main tables compare the unperturbed control reference; GEARS~\cite{roohani2024predicting} and GraphVCI~\cite{wu2022predicting}; scGPT~\cite{cui2024scgpt}, scFoundation~\cite{hao2024large}, GeneCompass~\cite{yang2024genecompass}, CellFM~\cite{zeng2025cellfm}, and CellVQ~\cite{wang2026cellvq}; and CPA~\cite{lotfollahi2023predicting}, STATE~\cite{adduri2025predicting}, CellFlow~\cite{klein2025cellflow}, PerturbDiff~\cite{yuan2026perturbdiff}, and scBIG~\cite{ruan2026scbig}. The control row repeats the unperturbed mean and provides a reference for the value of modeling an intervention. The methods shown in each table reflect the perturbation modality and available outputs for that setting. Scores use each method's available conditions; a dash marks an unavailable result. For \method, the evaluated conditions cover 383 of 384 RPE1 conditions and 2,050 of 2,053 transfer conditions.

\noindent\textbf{Metrics.}
Let $\mathcal{P}$ be the evaluated perturbation conditions, $\boldsymbol{\mu}_0$ the control mean, and $\boldsymbol{\mu}_p$ and $\widehat{\boldsymbol{\mu}}_p$ the observed and predicted means for condition $p$. Define $\Delta_p=\boldsymbol{\mu}_p-\boldsymbol{\mu}_0$ and $\widehat{\Delta}_p=\widehat{\boldsymbol{\mu}}_p-\boldsymbol{\mu}_0$. The response subset $\mathcal{D}_p$ contains the 20 genes with the largest observed absolute mean shifts and is distinct from the Wilcoxon-significant set used by DES.

\noindent\textbf{Response direction and identification.}
$\rho\Delta$ is the macro-average Pearson correlation between $\widehat{\Delta}_p$ and $\Delta_p$ over all genes; $\rho\Delta^{\mathrm{D}}$ applies the same calculation to $\mathcal{D}_p$. The comparison tables label these correlations Pearson $\Delta$ and Pearson $\Delta^{\mathrm{D}}$, respectively. The former tests whether the predicted response follows the observed full-gene pattern, whereas the latter focuses on the genes with the largest observed shifts. Directional accuracy on the response subset is
\begin{equation}
 \mathrm{ACC}\Delta^{\mathrm{D}}=\frac{1}{20|\mathcal{P}|}\sum_{p\in\mathcal{P}}\sum_{g\in\mathcal{D}_p}
 \mathbb{I}\!\left[\operatorname{sgn}(\widehat{\Delta}_{pg})=
 \operatorname{sgn}(\Delta_{pg})\right],
 \label{eq:acc_delta}
\end{equation}
DES is the Spearman correlation between predicted and observed log-fold changes over the protocol-defined Wilcoxon-significant genes. Together, these metrics assess correlation, sign agreement, and recovery of significant response structure.

\noindent\textbf{Condition-mean error.}
$L_2$, MSE, and MAE compare $\widehat{\boldsymbol{\mu}}_p$ with $\boldsymbol{\mu}_p$ over the complete gene universe and are macro-averaged over conditions. They quantify absolute expression reconstruction, which is useful alongside the change-from-control metrics: a model may reproduce the mean expression state yet miss the intervention response, or recover a response direction while shifting the overall mean.

\noindent\textbf{Mean-response similarity.}
Discriminative cosine compares the predicted and observed mean response directions relative to control within the distribution protocol. Higher values are better for $\rho\Delta$, $\rho\Delta^{\mathrm{D}}$, $\mathrm{ACC}\Delta^{\mathrm{D}}$, DES, and discriminative cosine; lower values are better for $L_2$, MSE, and MAE.

\noindent\textbf{Implementation details.}
Training follows Section~\ref{sec:training}; each setting has its own GRC scaffold, set cache, decoder calibration, and validation-selected flow checkpoint. JAX, Flax, and Optax implement the model. Genetic perturbations use their target representations, while RDKit supplies 2048-bit radius-2 Morgan fingerprints for ComboSciPlex without changing the gene encoder or conditional-flow interface. Experiments use two NVIDIA RTX PRO 6000 Blackwell Server Edition GPUs with 96 GB memory each, global pipeline seed 1, GRC seed 42, and cache-sampling seed 1.

\subsection{Full-Gene Genetic Perturbation Results}
\label{sec:genetic-results}
\noindent\textbf{Benchmark summary.}
Tables~\ref{tab:main-norman-additive}, \ref{tab:main-norman-holdout}, and~\ref{tab:main-replogle-holdout} report the Norman and RPE1 settings over all retained genes. Norman additive tests the ability to compose previously observed single-target responses, while the two target-holdout rows ask whether this ability carries over when the intervention gene itself was unseen during training. The RPE1 setting tests the same full-gene prediction objective under a different CRISPR screen and cell background. All entries are single-seed estimates; boldface and underlining mark the best and second-best prediction scores, with double-holdout MSE ties underlined at reported precision. Percentage changes below are relative changes calculated from unrounded scores.

\noindent\textbf{Unseen combinations and perturbation genes.}
On Norman additive, \method\ improves full-gene response correlation over scBIG by 4.4\% and correlation on the top 20 response genes by 9.9\%, while reducing MSE by 12.0\%. These changes indicate that the predicted expression shift follows the observed combination response more closely, including the genes with the largest effects. For held-out singles, full-gene response correlation is 0.7370 versus 0.7333 for scBIG, and MSE falls from 0.0034 to 0.0026. For held-out doubles, correlation reaches 0.8937 versus 0.8544, while MSE falls from 0.0028 to 0.0016. The clearer unseen-target correlation gain occurs for doubles; both splits improve mean-expression reconstruction. Figure~\ref{fig:norman-rho-delta} shows variation across individual perturbations.

\noindent\textbf{Cell-context dependence.}
On Replogle RPE1, \method\ improves full-gene response correlation over scBIG by 2.1\% and reduces MSE by 11.9\% (Table~\ref{tab:main-replogle-holdout}). Response recovery and absolute reconstruction improve together in this CRISPRi setting. The model also has the strongest displayed $L_2$ and discriminative-cosine scores, linking its mean-response gain to a closer predicted shift from the control population. These results extend the full-gene modeling approach from activating genes in K562 to inhibiting genes in RPE1, while the single-seed protocol leaves uncertainty about small ranking differences.

The UMAP examples in Figure~\ref{fig:cell-distribution-umap} compare the generated and observed cell populations rather than only their means. For Norman \textit{KLF1+MAP2K6}, \method\ follows the main elongated response region; for Replogle \textit{ATIC}, it reaches much of the observed region while assigning different densities to its subregions. These cases make the distinction between mean-response accuracy and the frequency of individual cell states visible. They also motivate evaluating population structure in addition to condition-level expression summaries.

\subsection{Cross-Context and Chemical Generalization}
\label{sec:cross-context-results}
\noindent\textbf{Transfer summary.}
The remaining two settings change the information available for generalization: cross-cell-line transfer changes the target cellular context, while ComboSciPlex changes the intervention description from a gene target to a chemical fingerprint. In strict cross-cell-line transfer, \method\ raises full-gene response correlation by 64.4\% and reduces MSE by 70.8\% relative to scGPT, the next-best method shown in Table~\ref{tab:secondary-transfer} on both metrics. This result is specific to the displayed K562 transfer protocol. The unchanged-control reference retains a lower MSE (0.0075 versus 0.0143) but cannot represent a perturbation-induced shift. The contrast separates basal-profile reconstruction from prediction of the intervention response.

On ComboSciPlex, \method\ improves correlation on the top 20 response genes by 8.8\% and reduces MSE by 17.9\% relative to PerturbDiff among the displayed methods in Figure~\ref{fig:drug-perturbation-metrics}. The corresponding full-gene response correlation and direction-accuracy panels show that the benefit spans complementary views of the drug response. Here the perturbation embedding is replaced by a Morgan fingerprint, whereas the full-gene encoder, population conditioning, and flow interface are retained. This result extends the framework to held-out chemicals without introducing a separate drug-specific prediction architecture.

\subsection{Diagnostic Experiments and Biological Programs}
\label{sec:interpretability}
The benchmark results show where \method\ predicts well; the following analyses test how its added components contribute and what the learned modules represent. Table~\ref{tab:diagnostics} reports matched-objective ablations, set-size sensitivity, functional coherence, and context-dependent assignments on Norman. Figure~\ref{fig:biology-programs} then tests whether the generated cells recover coordinated biological responses, including effects of double perturbations that differ from adding two single-target responses. Component and set-size experiments use the q-only configuration (Section~\ref{sec:softsub}).

\noindent\textbf{Component ablations and set size.}
Table~\ref{tab:diagnostics}(a) separates the contribution of Dynamic SoftSub from the two population paths. With SoftSub removed while Set-Cluster-Attention (SCA) and the source-set summary $s_0$ are retained, $\rho\Delta$ falls from 0.8521 to 0.8157 and MSE rises from 0.0015 to 0.001940. Adding expression-dependent local assignments therefore improves full-gene response correlation and mean reconstruction within the population-conditioned model. Shuffling the learned assignments also reduces $\rho\Delta$ to 0.8271, indicating that the gain depends on the learned assignment pattern. Removing SCA or $s_0$ separately gives smaller changes, while removing both reduces $\rho\Delta$ to 0.8344.

Increasing the source set from one to 32 cells raises $\rho\Delta$ by 33.2\% and approximately halves MSE (Table~\ref{tab:diagnostics}(b)). This is the clearest direct evidence that the observed control population supplies useful context beyond a single source cell. The default $N=128$ is selected by validation response correlation; $N=64$ gives the best test $\rho\Delta$ and MSE, and performance declines at $N=256$. Permuting cells in a fixed set changes latent outputs and metrics only at approximately $10^{-6}$, consistent with the intended permutation behavior.

\noindent\textbf{Functional refinement and context-dependent assignments.}
We next examine whether Dynamic SoftSub refines the fixed parent windows into gene groups with more coherent biological annotations. Learned assignments in 304 Norman parent windows are compared with 200 matched random partitions per window, preserving the gene set, effective slot sizes, and slot mass. Assignments are aggregated from 166 training conditions. Hard gain is the reduction in normalized annotation entropy based on dominant-slot labels. Soft gain is the increase in shared-annotation rate weighted by continuous gene-pair co-assignment, evaluated against within-window membership permutations. GO Biological Process and DoRothEA regulons~\cite{garcia2019dorothea} show positive gains; 59.9\% and 67.1\% of parent windows, respectively, exceed their matched hard-partition nulls (Table~\ref{tab:diagnostics}(c)). The two measures connect discrete annotation coherence with the continuous memberships used by the predictor.

We then measure how the learned grouping responds to cellular state. Relative to the training-control reference, assignment probabilities shift across expression contexts beyond within-condition resampling variation, measured by excess Jensen--Shannon divergence (Table~\ref{tab:diagnostics}(d)). Only about 1\% of dominant assignments change. Within each condition, however, gene-wise response amplitude and assignment divergence have mean Spearman correlations of 0.823, 0.858, and 0.836 in the training, validation, and test splits. Under IRF1+SET, HLA-B co-assignment with HLA-DRA/DPA1 falls from 0.894 to approximately 0.424; the corresponding observed shifts are 0.588 for HLA-B and 0.032/0.036 for HLA-DRA/DPA1. Under FOXA1+FOXA3, TFF3--TFF1 co-assignment falls from 0.960 to 0.577, alongside shifts of 0.379 and 0.079. These examples show that continuous memberships separate genes with unequal response amplitudes even when they occupy the same parent window.

\noindent\textbf{Biological program prediction.}
We score each program by averaging marker-gene responses standardized against training controls, with a standard-deviation floor of 0.25 and intervention targets excluded. The four ablation variants use 64 generated cells per condition; ranking and dominant-program analyses use 128. Across 32 test combinations and 13 programs, Dynamic SoftSub reduces program-response MAE from 0.0702 to 0.0579 relative to No SoftSub, which retains both population components. MAE is 17.5\% lower overall and lower in 12 of 13 programs (paired condition-bootstrap 95\% CI for the MAE difference, Dynamic SoftSub minus No SoftSub: $[-0.0221,-0.0023]$). The largest reductions occur for DNA replication (27.8\%) and the trefoil epithelial panel (24.8\%). MHC-II antigen presentation is the exception: its MAE increases from 0.0046 to 0.0088. The gains therefore concentrate on programs whose marker responses vary coherently across the test combinations, while the MHC-II panel remains sensitive to amplitude error. For 964 slots with gene weights fixed from training, perturbation-specific response ranking reaches NDCG@10 of 0.912, compared with 0.799 for the training-mean response and 0.740 for disjoint-target predictions (Figure~\ref{fig:biology-programs}(b)). Dynamic SoftSub thus improves both program-level response estimation and the localization of responsive gene groups.

\noindent\textbf{Combination-specific responses.}
Two Norman combinations illustrate nonadditive program behavior (Figure~\ref{fig:biology-programs}(d,e))~\cite{norman2019exploring}. In the CEBPE+PTPN12 validation combination, the observed myeloid program score falls from 0.988 under CEBPE alone to 0.344 under the double perturbation; addition predicts 0.948, whereas \method\ predicts 0.392. At the gene level, the model gives 0.106 MAE over the 12-gene panel, compared with 0.579 for addition, an 81.6\% reduction. In the PTPN12+UBASH3B test combination, the observed erythroid score rises to 0.485, compared with an additive expectation of 0.277; the model predicts 0.466. ALAS2, HBA1, HBA2, and HBG2 all show larger observed combination responses than their single-target additive values; for ALAS2, the observed, predicted, and additive responses are 1.857, 1.658, and 0.940, respectively. The model reduces the 18-gene panel MAE from 0.280 for addition to 0.100, a 64.3\% reduction. Across all 32 test combinations and 13 programs, the mean MAE is 0.0546 for \method\ and 0.0508 for addition, so the case-level gains identify specific nonadditive responses rather than a uniform advantage over addition.

\noindent\textbf{Dominant program recovery.}
Among 14 fixed marker panels, the predicted highest-scoring program matches the observed one in 27 of 32 test combinations (84.4\%), compared with 26 of 32 for additive composition. For the 15 combinations with literature-prior coverage, the predicted and observed leading panels agree in all 15 cases, compared with 13 of 15 for addition. IRF1 is linked to interferon-stimulated and MHC-I genes~\cite{abouelhassan2017properties}, while KLF1 is linked to erythroid transcription~\cite{gnanapragasam2025klf1}. The agreement is visible in IRF1+SET, whose leading panel is MHC-I, and KLF1+MAP2K6, whose leading panel is erythroid. Their amplitudes remain heterogeneous: predicted and observed MHC-I scores for IRF1+SET are 0.090 and 1.398, whereas erythroid scores for KLF1+MAP2K6 are 0.273 and 0.421. Program identity and response magnitude therefore provide complementary evidence about the generated state.

\section{Conclusion}
\label{sec:conclusion}

We presented \method\ for population-conditioned perturbation prediction across the full measured gene space. Dynamic SoftSub adapts local gene representations to each cell's expression, while Set-GCAE and a shared source summary incorporate control-population context into latent flow generation. Progressive training supports full-gene reconstruction across 6,539--19,451 genes. Evaluation on six genetic and chemical settings shows improved response correlation over scBIG on all three Norman protocols, with competitive performance on Replogle, cross-cell-line transfer, and held-out drugs. Component ablations and functional analyses support the contributions of adaptive local representations and population conditioning. The framework provides a basis for studying coordinated transcriptional responses across cellular states; future work can extend population modeling to rare states and more diverse cellular contexts.

\bibliographystyle{IEEEtran}
\bibliography{references}
\end{document}